\documentclass{article}
\usepackage[skins]{tcolorbox}  
\usepackage{graphicx}  
\usepackage{xcolor}    
\usepackage{amsmath}   
\usepackage{amssymb}   
\usepackage{hyperref}
\usepackage{multirow}
\usepackage{float}
\usepackage{hyperref}
\usepackage{caption}

\tcbset{
    colback=green!10!gray!5,   
    colframe=green!50!gray!50, 
    arc=3mm,                   
    width=0.9\textwidth,       
    boxrule=0pt,               
    boxsep=10pt,               
    top=20pt, bottom=5pt,    
    enhanced,                  
    center                    
}

\author{Seyedeh Sahar Taheri Otaghsara, Reza Rahmanzadeh}
\date{}  

 \usepackage[a4paper, top=2cm, bottom=2.5cm, left=2.5cm, right=2.5cm]{geometry}

\begin{document}

\vspace*{-0.5cm}  

\begin{center} 
    \begin{tcolorbox}
    
        \vspace{-0.5cm} 
        
       \hspace{2.5cm}{\scriptsize Towards Foundation Models for Medical Image Segmentation}\

 \vspace{0.3cm} 
 
          \begin{center}
    \textbf{\large Multi-encoder nnU-Net outperforms transformer models with self-supervised pretraining}
\end{center}

        \vspace{1cm}  
        \textbf{Seyedeh Sahar Taheri Otaghsara} DDS,   
        \textbf{Reza Rahmanzadeh}  MD, PhD
        
 \vspace{0.4cm} 
        \raggedright {\small \textcolor{blue}{\texttt{ AI Lab, UltraAI}}} 

        \vspace{0.5cm} 

        This study addresses the essential task of medical image segmentation, which involves the automatic identification and delineation of anatomical structures and pathological regions in medical images. Accurate segmentation is crucial in radiology, as it aids in the precise localization of abnormalities such as tumors, thereby enabling effective diagnosis, treatment planning, and monitoring of disease progression. Specifically, the size, shape, and location of tumors can significantly influence clinical decision-making and therapeutic strategies, making accurate segmentation a key component of radiological workflows. However, challenges posed by variations in MRI modalities, image artifacts, and the scarcity of labeled data complicate the segmentation task and impact the performance of traditional models.

To overcome these limitations, we propose a novel self-supervised learning Multi-encoder nnU-Net architecture designed to process multiple MRI modalities independently through separate encoders. This approach allows the model to capture modality-specific features before fusing them for the final segmentation, thus improving accuracy. Our Multi-encoder nnU-Net demonstrates exceptional performance, achieving a Dice Similarity Coefficient (DSC) of 93.72\%, which surpasses that of other models such as vanilla nnU-Net, SegResNet, and Swin UNETR. By leveraging the unique information provided by each modality, the model enhances segmentation tasks, particularly in scenarios with limited annotated data. Evaluations highlight the effectiveness of this architecture in improving tumor segmentation outcomes.

\vspace{0.5cm} 
\raggedright {\small March 2025}  

 \vspace{0.2cm} 
        \raggedright {\small Correspondence: \textcolor{blue}{\texttt{reza@theultra.ai}}} 

   \vspace{0.3cm} 
\hspace{0.3cm} 
\hfill 
\raisebox{-0.1cm}{\includegraphics[width=2cm]{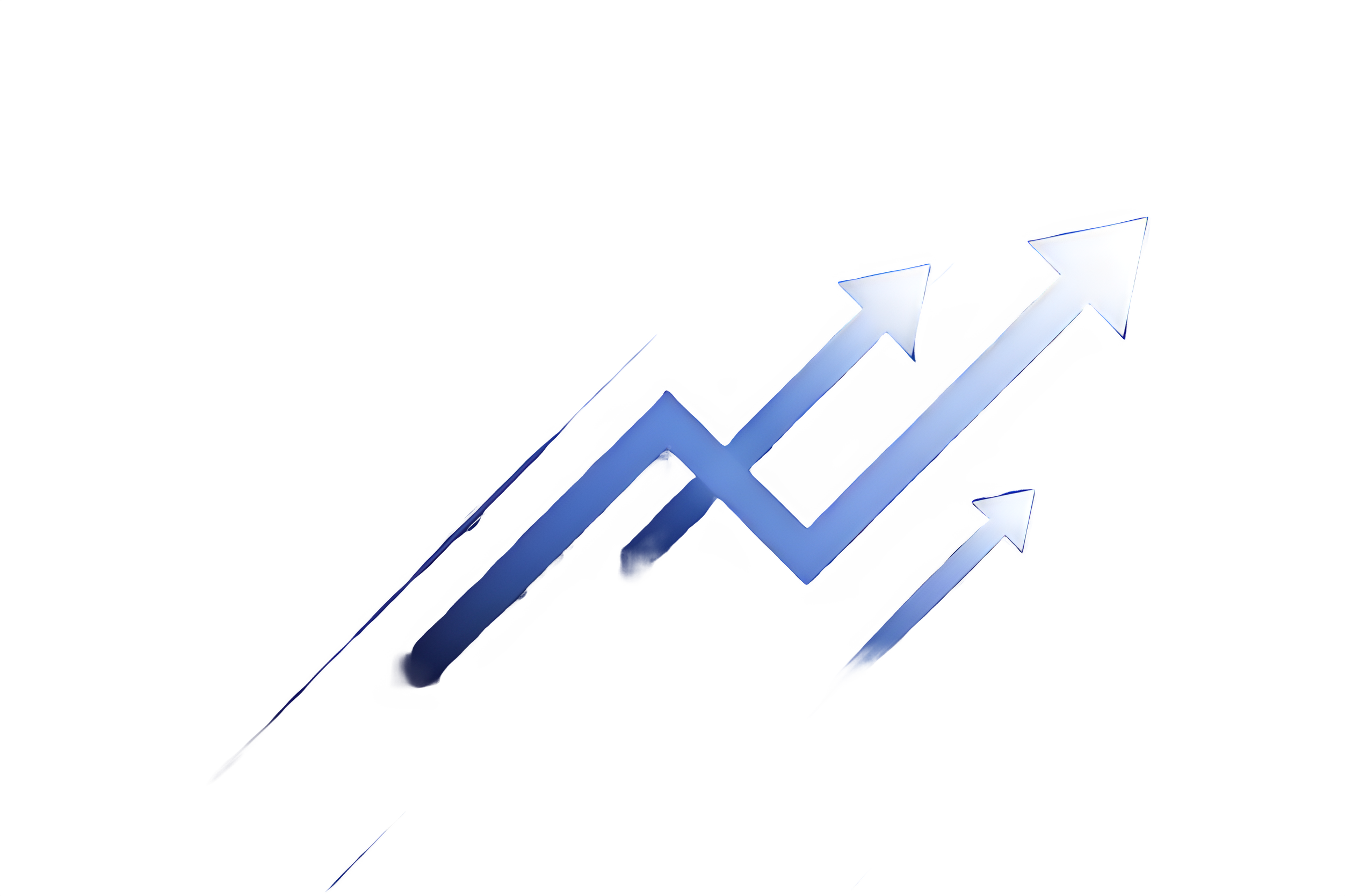}}  
\hspace{-0.5cm}  
\includegraphics[width=3cm]{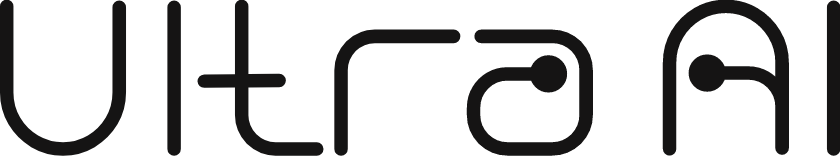}  
       
    \end{tcolorbox}
\end{center}

\vspace{0.2cm} 

\section{Introduction}
The integration of segmentation and detection technologies into clinical practice represents a transformative shift in medical imaging \cite{litjens2017survey}\cite{shen2017}. By automating the identification and delineation of anatomical structures and pathological regions, these advanced methodologies significantly enhance diagnostic accuracy and clinical decision-making \cite{he2019practical}\cite{visser2019inter}. Accurate segmentation not only streamlines the workflow for healthcare professionals but also fosters improved inter-rater reliability—the consistency of diagnostic interpretations among different clinicians \cite{sharei2018navigation}. This is particularly crucial in fields such as oncology, where precise localization of tumors can dictate treatment pathways and prognostic assessments \cite{tripathi2024survey}.

\vspace{0.2cm}
Despite the promising advancements in segmentation models, the field continues to face several challenges. One primary obstacle stems from the inherent variations in MRI modalities, which can influence the quality and interpretability of images \cite{paschali2018generalizability}. Variations in acquisition techniques—ranging from the choice of scanners to the application of specific MR sequences and reconstruction algorithms—introduce significant discrepancies in image characteristics \cite{alzubaidi2021novel}. Additionally, the presence of MRI artifacts, such as motion-induced distortions and magnetic field inhomogeneities, further complicates the segmentation task, necessitating sophisticated, adaptable models that can maintain robustness across diverse imaging scenarios \cite{chen2020simple}\cite{dosovitskiy2014}\cite{js2025brain}.

\vspace{0.2cm}
In this context, a critical limitation of current segmentation models is their generalizability, particularly their susceptibility to out-of-distribution errors \cite{bakas2017advancing}. This issue is exacerbated by the scarcity of annotated medical datasets, which are often prohibitively expensive and time-consuming to compile due to the requirement for expert annotation \cite{antonelli2022medical}\cite{clark2013cancer}. As a result, many existing models are trained on limited datasets, leading to a tendency to overfit, demonstrating high accuracy on similar data but poor performance when applied to new, unseen images \cite{ting2023multimodal}. This lack of resilience when confronted with variations not represented in their training data highlights the pressing need for models that can learn effectively from less labeled data \cite{vaswani2017}.

\vspace{0.2cm}
To address these limitations, self-supervised learning (SSL) has emerged as a compelling solution, offering innovative techniques to leverage vast amounts of unlabeled data \cite{etelvina2025harnessing}\cite{yadav2024leveraging}. SSL can be classified into several approaches, including masked self-supervised learning (masked SSL), which involves predicting masked portions of the data \cite{xu2024self}, and contrastive self-supervised learning (contrastive SSL), which focuses on learning representations by contrasting similar and dissimilar data samples \cite{oord2018}\cite{ilani2025t1}. Additionally, few-shot learning (FSL) \cite{sowjanya2025enhancing} and semi-supervised learning are vital techniques that complement SSL \cite{liu2025glioma}. FSL enables models to learn from only a handful of labeled examples, making it particularly useful in domains with limited annotated data \cite{thakkar2018part}. In contrast, semi-supervised learning combines a small amount of labeled data with a large amount of unlabeled data, allowing models to improve their performance by leveraging the structure of the unlabeled dataset alongside the labeled instances \cite{luo2021semi}.

\vspace{0.2cm}
 The high diversity of medical imaging segmentation tasks underscores the necessity for foundation models capable of generalizing across a multitude of applications \cite{zhang2024challenges}. Notable existing medical datasets that facilitate research in this domain include the Brain Tumor Segmentation (BraTS) challenge, which focuses on the segmentation of brain tumors in MRI scans \cite{de20242024}, the Medical Segmentation Decathlon (MSD) \cite{antonelli2022medical}, and the ATLAS challenge \cite{quinton2023tumour}. Other significant datasets include the ISLES (Ischemic Stroke Lesion Segmentation) challenge \cite{hernandez2022isles} and several MS datasets that provide critical benchmarks for evaluating model performance \cite{commowick2021multiple}. Furthermore, large-scale medical image datasets such as CheXpert \cite{irvin2019chexpert}, MIMIC-CXR \cite{johnson2019mimic}, and RadImageNet \cite{mei2022radimagenet} are emerging as valuable resources for training and validating models in the radiology domain. These datasets not only aid in training and validating models but also promote collaboration and knowledge sharing within the medical imaging community \cite{adewole2023brain}. Additionally, general computer vision datasets, such as ImageNet, play a crucial role in pre-training models for various applications, providing a rich source of labeled data that can enhance transfer learning capabilities \cite{deng2009imagenet}.

\vspace{0.2cm}
An important aspect of improving segmentation in clinical settings is the consideration of multimodal imaging, where different MRI modalities are utilized to capture unique biological information about a patient \cite{isensee2021b}\cite{isensee2021c}\cite{lakshmi2025explainable}. Multimodal segmentation tools, which integrate data from various imaging techniques—such as T1-weighted MRI and diffusion-weighted MRI—offer a more comprehensive understanding of complex medical conditions \cite{kamnitsas2018}.
However, a significant limitation arises when models input all modalities into a single encoder. This approach can constrain the model's ability to learn modality-specific patterns, which are critical for accurately interpreting the distinct biological targets represented by each modality \cite{kamnitsas2015}\cite{kamnitsas2017}. To address this challenge, we propose a novel modified nnU-Net architecture that incorporates separate encoders for each MRI modality. By learning high-level features independently before merging the knowledge acquired from distinct modalities, our model aims to enhance overall segmentation accuracy, ultimately leading to improved clinical outcomes\cite{dorfner2025review}\cite{long2015}.

\vspace{0.2cm}
The challenge of limited labeled data in medical imaging is pervasive, often hindering the development and deployment of effective machine learning models \cite{luu2021}\cite{xu2025generalizable}. As explained earlier, to mitigate this issue, SSL has gained traction as a viable approach, enabling models to learn from vast amounts of unlabeled data while requiring minimal labeled examples \cite{mckinley2019}\cite{ronneberger2015}\cite{cox2024brainsegfounder}. SSL techniques can help models develop a foundational understanding of the data, which can then be fine-tuned for specific downstream tasks with limited labeled data \cite{etelvina2025harnessing}. This process involves training the model on a related task or domain and subsequently refining its parameters to adapt to the nuances of the target task, thereby improving performance despite the initial lack of annotated data. Transfer learning also plays a crucial role in this context, allowing models pre-trained on large datasets to be adapted for specific medical imaging challenges \cite{mei2022}. By leveraging existing knowledge, transfer learning can significantly reduce the data requirements for effective model training, enabling more robust and generalizable segmentation outcomes \cite{kora2022transfer}.

\vspace{0.2cm}
In this paper, we present a comprehensive comparative analysis of various architectural models, including state-of-the-art U-Net  models \cite{mallampati2023brain} such as vanilla nnU-Net \cite{isensee2021b} and Multi-encoder nnU-Net \cite{xue2025mmefu}, versus Transformer models \cite{hatamizadeh2021b}. Additionally, we explore different training strategies, specifically those involving self-supervised learning, to assess their impact on model performance  \cite{ma2023}\cite{song2025rehrseg}\cite{fang2021self}. By investigating these models and strategies, we aim to elucidate the potential of advanced architectures and learning paradigms in overcoming the current limitations in medical image segmentation, ultimately advancing the state of the art and improving clinical outcomes.

\section{Related Works}
In the previous BraTS challenges, ensembles of U-Net shaped architectures have achieved promising results for multi-modal brain tumor segmentation. Kamnitsas et al. \cite{kamnitsas2018}, the winners of the 2017 BraTS challenge, introduced the ensemble of multiple models and architectures (EMMA), which incorporates 3D convolutional networks such as DeepMedic \cite{kamnitsas2015}\cite{kamnitsas2017}, FCN \cite{long2015}, and U-Net \cite{ronneberger2015}\cite{saleh2025traditional}. EMMA leverages the strengths of various models to reduce the influence of meta-parameters and mitigate overfitting, offering more robust segmentation results for brain tumors.

For the 2020 and 2021 BraTS challenges \cite{isensee2021c}\cite{baid2021}, the winning teams proposed the nnU-Net \cite{isensee2021b}, a self-configuring U-Net-based architecture, as a baseline. They implemented several BraTS-specific optimizations, demonstrating its adaptability and effectiveness for tumor segmentation tasks.

In 2022, the BraTS challenge winners \cite{zeineldin2022} achieved the best performance using an ensemble of three distinct architectures: DeepSeg \cite{zeineldin2020}, an enhanced version of nnU-Net \cite{luu2021}, and DeepSCAN \cite{mckinley2019}. The ensemble method was built using the Simultaneous Truth and Performance Level Estimation (STAPLE) technique.

Similarly, the 2023 BraTS-Africa challenge \cite{adewole2023brain} employed the STAPLE ensemble of three models to generate ground truth segmentations for glioma patients from sub-Saharan Africa.

\vspace{0.2cm}
The nnU-Net framework \cite{isensee2021b}, a fully automated, self-configuring system, has been widely used as a baseline for brain tumor segmentation, particularly in its 3D full-resolution variant, which has been applied without any further configuration changes.

\vspace{0.2cm}
Hatamizadeh et al. \cite{hatamizadeh2021b} proposed the UNETR architecture in which a Vision Transformer (ViT)-based encoder, which directly utilizes 3D input patches, is connected to a CNN-based decoder. UNETR has shown promising results for brain tumor segmentation using the MSD dataset \cite{antonelli2022medical}.

\vspace{0.2cm}
The Swin UNETR \cite{hatamizadeh2021} is another significant contribution, where the traditional convolutional encoder in U-Net is replaced with Swin Transformer blocks. This allows the model to capture long-range dependencies and global contextual information, which fully convolutional networks struggle to represent. The Swin Transformer utilizes shifted windows to process high-resolution images efficiently, making it particularly suitable for datasets like BraTS, where large image sizes are common \cite{liu2021}\cite{zhou2021}.

\vspace{0.2cm}
In medical language processing, models such as MI-Zero \cite{lu2023} and BioViL-T \cite{bannur2023} have used contrastive learning to push forward representational analysis and zero-shot transfer learning for medical image recognition. These models use image-text pairs to refine segmentation by pulling similar pairs closer in the latent space while pushing dissimilar pairs apart, contributing to advances in histopathology research and multimodal image analysis. However, they depend on the availability of text-based prompts accompanying the training images \cite{tiu2022}.

\vspace{0.2cm}
Despite the progress made with convolutional and transformer-based architectures, medical image segmentation has yet to fully benefit from the recent advances in natural image analysis and language processing. Models such as the Segment Anything Model (SAM) \cite{kirillov2023}\cite{diana2025gbt} and LLaMA \cite{touvron2023} have shown impressive results in natural image segmentation tasks, but their adaptation to medical imaging remains underexplored.
Following SAM's success in few-shot segmentation of natural images, several recent works have focused on adapting SAM to medical image segmentation. MedSAM \cite{ma2023}, MedLSAM \cite{lei2023} and SAM-Med2D \cite{cheng2023} modify SAM’s architecture to improve its performance on medical imaging tasks, bridging the gap between SAM's generalizability to real-world images and the challenges posed by medical datasets.

\section{Dataset}
Recent efforts have focused on the development of extensive medical datasets \cite{mei2022}\cite{clark2013}\cite{bycroft2018}. In this study, we specifically utilized two datasets: the UK Biobank and the BraTS dataset.

\subsection{BraTS Dataset}
The BraTS dataset features a retrospective collection of multi-institutional, multi-parametric MRI scans of brain tumors \cite{baid2021}. These scans were obtained under standard clinical conditions but with varying equipment and imaging protocols, resulting in a diverse range of image quality that mirrors different clinical practices across institutions. To be included in the dataset, participants needed a pathologically confirmed diagnosis and available MGMT promoter methylation status. Expert neuroradiologists approved the ground truth annotations for each tumor sub-region, while MGMT methylation status was determined through laboratory assessments of surgical brain tumor specimens.

\vspace{0.5cm}
\textbf{Imaging Data Description}
The MRI scans used in the BraTS 2021 challenge consist of four types: a: native (T1), b: post-contrast T1-weighted (T1Gd, using gadolinium), c: T2-weighted (T2), and d: T2 Fluid Attenuated Inversion Recovery (T2-FLAIR) volumes. These scans were acquired using various protocols and scanners from multiple institutions.
All BraTS MRI scans underwent standardized pre-processing, which involved converting DICOM files to the NIfTI file format \cite{li2016}, co-registering them to a consistent anatomical template, resampling to a uniform isotropic resolution of 1 mm³, and performing skull stripping.
The imaging volumes were segmented using the STAPLE \cite{warfield2004} fusion of the top-performing BraTS algorithms, including nnU-Net \cite{isensee2021b}, DeepScan \cite{mckinley2019}, and DeepMedic \cite{kamnitsas2015}\cite{kamnitsas2017}. These fused labels were manually refined by volunteer neuroradiology experts with varying ranks and experience, adhering to a clearly defined annotation protocol. The final annotations were approved by board-certified attending neuroradiologists with over 15 years of experience in glioma work. The annotated tumor sub-regions are based on known features visible to trained radiologists (VASARI features) and include the Gd-enhancing tumor, peritumoral edematous/invaded tissue, and the necrotic tumor core.

\subsection{UK Biobank (UKB)}
We employed T1-weighted (T1w) and T2-weighted Fluid Attenuation Inversion Recovery (T2-FLAIR) images sourced from the UK Biobank (UKB) dataset \cite{littlejohns2020}. Collected since 2014 and preprocessed by the UKB, these images were part of a detailed 35-minute protocol that captured various brain imaging modalities, including T1w and T2-FLAIR structural MRI. Between 2014 and 2022, neuroimaging data were obtained from 44,172 participants. The raw T1w structural volumes underwent processing using a pipeline by UK Biobank researchers, largely relying on FSL and FreeSurfer tools \cite{woolrich2009}. T2-FLAIR images were co-registered with their corresponding T1 images.

DICOM files were converted to NIfTI format using dcm2niix \cite{li2016} and transferred to the MNI152 space using FNIRT. From the pool of 44,172 participants, 43,369 had available T1-weighted (T1w) and T2-FLAIR images. For creating 3D foundational models in neuroimaging, we focused on participants with a significant number of slices in both MRI modalities. This approach narrowed our dataset to 41,000 participants, yielding a total of 82,000 imaging volumes.

\vspace{0.5cm}
\textbf{Pre-processing}
Additional pre-processing, including z-score normalization and image augmentation, was performed on both datasets following the nnU-Net pipeline.

\section{Comparison models}
\subsection{U-Net architecture 
}
To comprehensively investigate U-Net performance in medical image segmentation task, we included four different U-Net based models in our comparison: custom U-Net (SegResNet), vanilla nnU-Net, Multi-encoder nnU-Net with and without pretraining.

\subsubsection{Multi-encoder nnU-Net}
\begin{figure}[H]  
    \centering
    \includegraphics[width=0.9\textwidth]{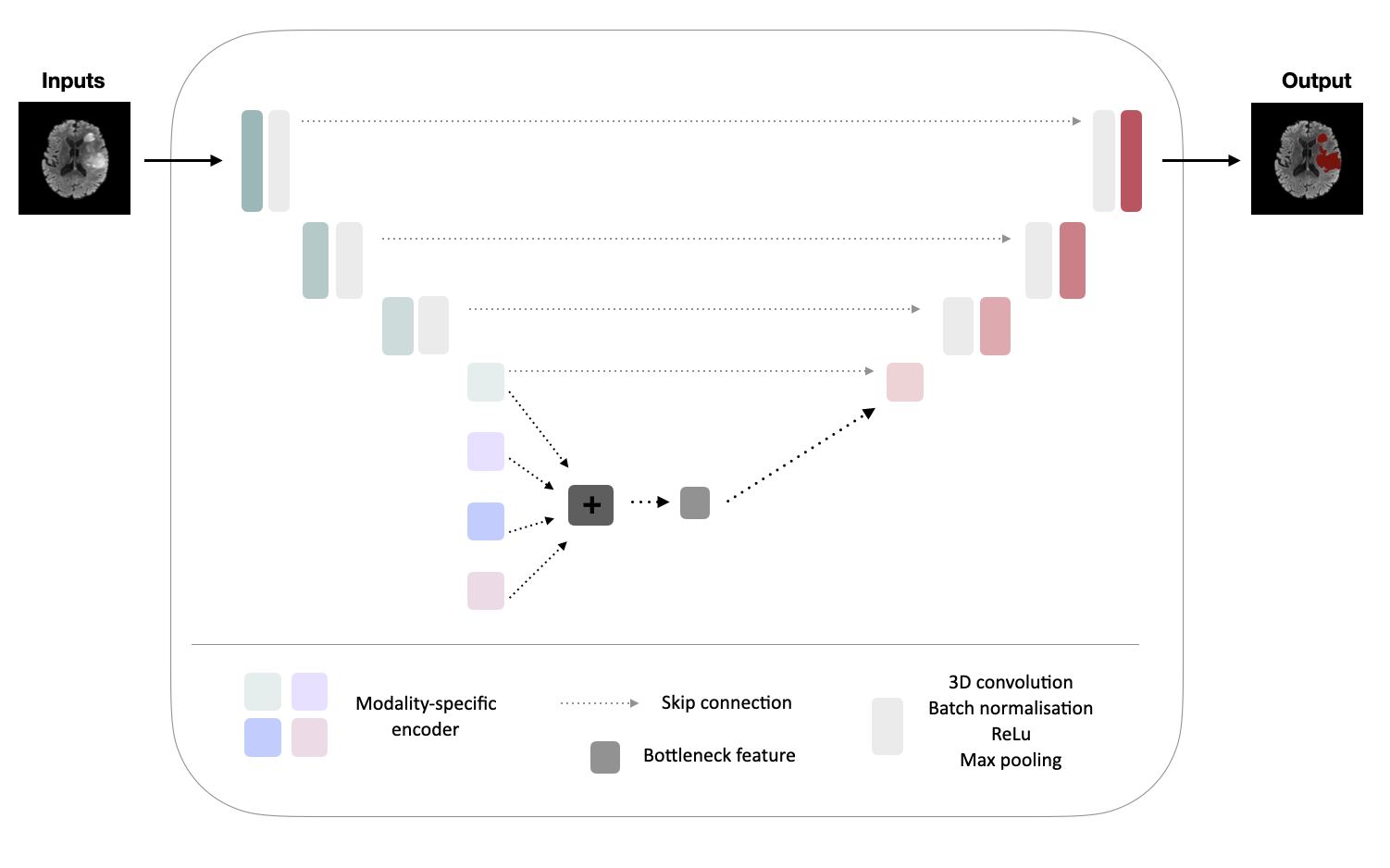}
    \caption {\textbf{ Overview of the Multi-encoder nnU-Net architecture.}  Each MRI modality is fed into a separate encoder, allowing for specialized feature extraction tailored to the unique characteristics of each modality. At the bottleneck layer, the encoded representations from all modalities are combined, integrating diverse information for comprehensive feature representation. This combined representation is then passed through a shared decoder, which generates a lesion mask that delineates all pathologies present across the input MRI modalities.}
\label{figure-1}

\end{figure}

Our approach centers around the nnU-Net framework \cite{isensee2021b}, which serves as the foundational architecture for segmentation. In our model, we implement separate encoders tailored for different imaging modalities, while a common decoder is utilized across the board (Figure \ref{figure-1}). Each input image is directed through its respective modality-specific encoder, and then the unified decoder produces anomaly segmentations. The segmentation task employs the following loss function consisting of two components:

\begin{equation}
\mathcal{L} = \lambda_1 \cdot \mathcal{L}_{\text{Dice}}(s, \hat{s}) + \lambda_2 \cdot \mathcal{L}_{\text{CE}}(s, \hat{s})
\end{equation}
\vspace{0.1cm}
where:

- \( \mathcal{L}_{\text{Dice}}(s, \hat{s}) \) is the Dice loss, which maximizes the overlap between the predicted and actual segmentation maps, controlled by the weight \( \lambda_1 \).

- \( \mathcal{L}_{\text{CE}}(s, \hat{s}) \) is the cross-entropy loss, which penalizes incorrect pixel predictions, improving the alignment of the predicted map with the true segmentation, controlled by \( \lambda_2 \).

Here, \( s \) acts as the supervisory label, and \( \hat{s} \) is the anticipated binary mask. The coefficient \( \lambda_1 \) pertains to the Dice loss, while \( \lambda_2 \) pertains to the cross-entropy loss.

\textbf{Training and Implementation Details}
During training, we establish the following hyperparameters:

\begin{itemize}
    \item \textbf{Global batch size}: 2
    \item \textbf{Input patch size}: \( (96, 112, 80) \)
    \item \textbf{Learning rate scheduler}: Polynomial decay with:
    \begin{equation}
        \eta_t = \eta_0 \times (1 - \frac{t}{T})^{0.9}
    \end{equation}
    where:
    \begin{itemize}
        \item \( \eta_t \) is the learning rate at epoch \( t \).
        \item \( \eta_0 = 10^{-2} \) is the initial learning rate.
        \item \( T \) is the total number of epochs.
    \end{itemize}
    
    \item \textbf{Optimizer}: Stochastic Gradient Descent (SGD) with:
    \begin{itemize}
        \item Weight decay: \( 3 \times 10^{-5} \)
        \item Momentum: 0.95
    \end{itemize}
    
    \item \textbf{Maximum training epochs}: 500
\end{itemize}
The model's encoder and decoder backbone, data preprocessing, and augmentation strategies adhere to the nnU-Net \cite{isensee2021b} framework.

\subsection{Vision transformer architecture }
To comprehensively investigate vision transformer performance in medical image segmentation task, we included different transformer based models in our comparison: Swin UNETR with and without pretraining.

\subsubsection{Swin UNETR}
\begin{figure}[H]  
    \centering
    \includegraphics[width=0.9\textwidth]{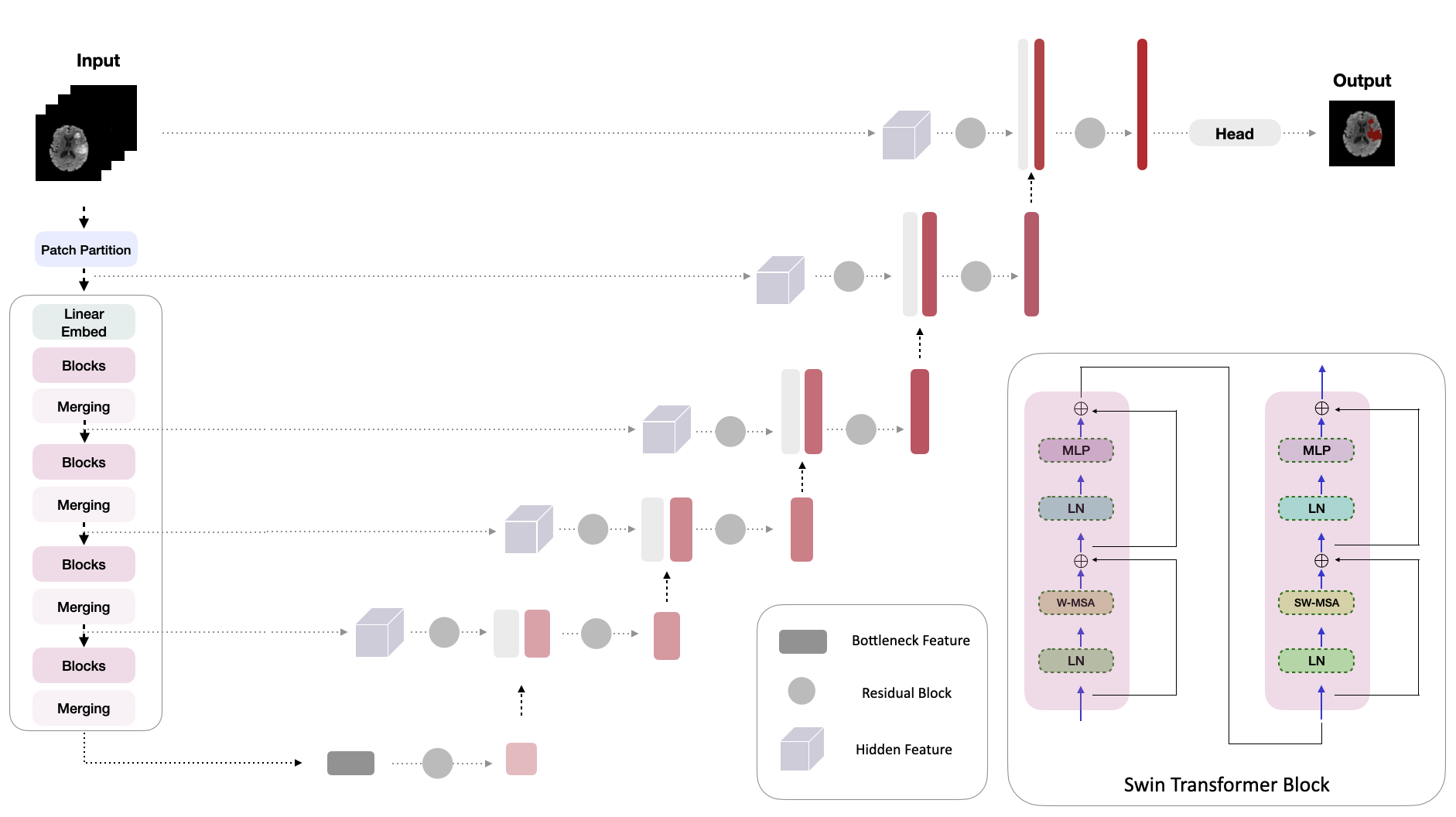}
    \caption{\textbf{Overview of the Swin UNETR Architecture.} The model processes 3D multi-modal MRI images with 4 channels as input. It segments the input into non-overlapping patches and utilizes a patch partition layer to create windows of a specific size for computing self-attention. The Swin transformer's encoded feature representations are transmitted to a CNN decoder via skip connections at multiple resolutions. The resulting segmentation output consists of 3 channels, each representing the ET, WT, and TC sub-regions. Finally, these three masks are binarized and combined to produce the final lesion mask.}
    \label{figure-2}  
\end{figure}

The Swin UNETR (Swin Transformer-based U-Net with Residual Connections) integrates the Swin Transformer for hierarchical feature extraction with a UNETR-style decoder to generate high-precision segmentation maps \cite{hatamizadeh2021}. It leverages hierarchical self-attention for multi-scale feature representation, while the skip connections in the decoder help retain spatial information for more accurate segmentation (Figure \ref{figure-2}). We employ the soft Dice loss function \cite{myronenko2019robust}, calculated voxel-wise as follows:

\[
L(G; Y) = 1 - \frac{2}{J} \sum_{j=1}^{J} \frac{\sum_{i=1}^{I} G_{i,j} Y_{i,j}}{\sum_{i=1}^{I} G_{i,j}^2 + \sum_{i=1}^{I} Y_{i,j}^2}
\]
where:

- \( I \) represents the number of voxels,  

- \( J \) denotes the number of classes,  

- \( Y_{i,j} \) corresponds to the predicted probability for class \( j \) at voxel \( i \), 

and  

- \( G_{i,j} \) is the one-hot encoded ground truth for class \( j \) at voxel \( i \).

\vspace{0.5cm}

\textbf{Training and Implementation Details}

During training, we establish the following hyperparameters:

\begin{itemize}
    \item \textbf{Encoder Backbone}: Swin Transformer with hierarchical feature learning.
    \item \textbf{Decoder}: Skip connections and upsampling layers for spatial preservation.
    \item \textbf{Optimizer}: AdamW with weight decay.
    \item \textbf{Learning Rate Scheduler}: Cosine Annealing.
    \item \textbf{Batch Size}: 4
    \item \textbf{Patch Size}: \( 96 \times 96 \times 96 \)
    \item \textbf{Number of Training Epochs}: 500
\end{itemize}

\subsection{Self-supervised learning (SSL) pretraining strategy
}

The pretraining strategy includes two separate stages: (1) Pretraining on the UK Biobank (UKB) and (2) Pretraining on the BraTS Dataset.

Following the methodology from \cite{Tang2022}, the first stage of pretraining involves self-supervised learning using a large, unlabeled dataset of images from the UKB dataset. We utilize 3D volumetric images for this pretraining process. The input MRI modalities are randomly cropped into sub-volumes, followed by image augmentation.

Following the approach outlined in \cite{yadav2024leveraging}, the pretraining of the Swin UNETR encoder is carried out using three unique proxy tasks that function as self-supervised fine-tuning methods: masked volume inpainting, 3D image rotation, and contrastive coding.

In the second phase, the models initially pretrained on the UKB dataset underwent additional pretraining through transfer learning on the Brain Tumor Segmentation (BraTS) dataset.

After completing the pretraining stages, the Multi-encoder nnU-Net and Swin UNTER models described above were fine-tuned using the BraTS dataset.

\subsection{TransBTS}

TransBTS (Transformer-Based Brain Tumor Segmentation) \cite{wang2021transbtsmultimodalbraintumor} combines CNNs with vision transformers (ViTs) for brain tumor segmentation. It uses a CNN encoder for feature extraction, a transformer bottleneck for global context modeling, and a CNN decoder for segmentation, enhancing tumor boundary delineation and overall segmentation performance.
\subsection{SegResNet}
It is a network designed for semantic segmentation, particularly aimed at segmenting tumor subregions in 3D MRIs \cite{myronenko20183dmribraintumor}. It uses an encoder-decoder framework and includes a variational auto-encoder branch to reconstruct the input image. This branch helps to regularize the shared decoder and adds constraints to its layers, which is especially beneficial given the limited training dataset size. The encoder is composed of ResNet blocks, each containing two convolutional layers with normalization and ReLU activation, followed by additive identity skip connections. Group Normalization is used for normalization. The decoder has a structure similar to the encoder's but with one block for each spatial level. This method secured the first position in the BraTS 2018 challenge.

\vspace{0.5cm}

\section{Metrics}
To evaluate the segmentation task, we employ a range of metrics for comparison. The performance is measured using the Dice similarity coefficient, accuracy, sensitivity, specificity, and precision. 

\subsection{DSC} The Dice Similarity Coefficient (DSC) is a conventional metric for segmentation that quantifies the overlap between the predicted output \( P \) and the actual ground truth \( G \), and is formally defined as follows:

\begin{equation}
\text{DSC} = \frac{2 |P \cap G|}{|P| + |G|}
\end{equation}

where:
\begin{itemize}
    \item \( P \) denotes the segmentation predicted by the model,
    \item \( G \) refers to the actual ground truth segmentation,
    \item \( |P| \) represents the size (or count of pixels/voxels) of the predicted segmentation,
    \item \( |G| \) indicates the size (or count of pixels/voxels) of the ground truth segmentation,
    \item \( |P \cap G| \) is the overlap or intersection between the predicted and actual ground truth segmentations.
\end{itemize}

\subsection{ACC}  Accuracy (ACC) is a standard metric used to evaluate the proportion of correct predictions made by the model. It is defined as the ratio of the number of correct predictions (both true positives and true negatives) to the total number of predictions, given by:

\[
\text{ACC} = \frac{TP + TN}{TP + TN + FP + FN}
\]

where:
\begin{itemize}
    \item \( TP \) = True Positives
    \item \( TN \) = True Negatives
    \item \( FP \) = False Positives
    \item \( FN \) = False Negatives
\end{itemize}

\subsection{SE} Sensitivity (SE) is a metric used in segmentation to gauge the model's effectiveness in accurately identifying patients with the disease. It is defined as follows:

\begin{equation}
\text{SE} = \frac{|P \cap G|}{|G|}
\end{equation}

where:
\begin{itemize}
    \item \( P \) stands for the segmentation predicted by the model,
    \item \( G \) signifies the actual ground truth segmentation,
    \item \( |P \cap G| \) refers to the overlap between the predicted and actual segmentations, representing the true positives,
    \item \( |G - P| \) denotes the portion of the ground truth segmentation that the model failed to predict, accounting for the false negatives.
\end{itemize}

\subsection{SP} Specificity (SP) is a segmentation metric that measures the model’s ability to correctly identify the negative cases. It is defined as the ratio of correctly identified negatives to the total number of actual negatives, which is defined as:
\[
\text{SP} = \frac{|P^c \cap G^c|}{|G^c|}
\]

Where:
\begin{itemize}
    \item \( P^c \) represents the complement of the predicted segmentation (the predicted negatives),
    \item \( G^c \) represents the complement of the ground truth segmentation (the actual negatives),
    \item \( |P^c \cap G^c| \) is the intersection of the predicted and ground truth negative segmentations (true negatives).
\end{itemize}

\subsection{PRE}  Precision (PRE) is a metric for segmentation that evaluates the likelihood of making correct predictions. It is defined as:

\begin{equation}
\text{PRE} = \frac{|P \cap G|}{|P|}
\end{equation}

where:
\begin{itemize}
    \item \( P \) represents the predicted segmentation,
    \item \( G \) represents the ground truth segmentation,
    \item \( |P \cap G| \) is the area (or number of pixels/voxels) of overlap between the predicted and ground truth segmentations (true positives),
    \item \( |P - G| \) is the area of the predicted segmentation that does not overlap with the ground truth (false positives).
\end{itemize}

\section{Results}
Table \ref{tab:model_comparison} presents a comparative analysis of our best-performing model, SSL Multi-encoder nnU-Net, against other state-of-the-art (SOTA) models in the BraTS challenge, including Swin UNETR \cite{hatamizadeh2021}, nnU-Net \cite{isensee2021b}, TransBTS \cite{wenxuan2021}, and SegResNet \cite{monai2024}.

\vspace{0.2cm}
SegResNet and nnU-Net have been among the winning methodologies in previous BraTS challenges, while TransBTS is a vision transformer-based approach specifically designed for brain tumor segmentation. To thoroughly assess the effectiveness of our proposed model, we evaluated its performance against these benchmark architectures.
The results demonstrate that the SSL Multi-encoder nnU-Net consistently outperforms all competing approaches across multiple evaluation metrics.

\vspace{0.2cm}
In terms of the average DSC, the SSL Multi-encoder nnU-Net achieved the highest score of 93.87\%, outperforming both nnU-Net (90.89\%) and SegResNet (92.00\%). SSL Pretraining improved model performance for both U-Net and transformer model architectures: SSL Multi-encoder nnU-Net (DSC 93.72) vs SL Multi-encoder nnU-Net (DSC 92.04) and SSL Swin UNETR (DSC 92.80) vs SL Swin UNETR (DSC 91.80). Furthermore, Multi-encoder nnU-Net outperformed Swin UNETR for both SL (92.04 vs 91.80) and SSL training (93.82 vs 92.80) strategies. 

\vspace{0.2cm}
Similarly, our SSL Multi-encoder nnU-Net model exhibited the highest performance in surpassing existing SOTA models across other key metrics, including ACC, SP, and PRE, demonstrating its potential as a powerful tool for brain tumor segmentation in clinical applications.

\begin{table}[h]
    \centering
    \renewcommand{\arraystretch}{1.3}
    \setlength{\tabcolsep}{4pt} 
    \scriptsize 
    \begin{tabular}{lccccc}
        \hline
        \scriptsize Methods & \scriptsize Av. DSC (\%) & \scriptsize Accuracy & \scriptsize Sensitivity & \scriptsize Specificity & \scriptsize Precision \\ 
        \hline
        \scriptsize SSL Multi-encoder nnU-Net & \scriptsize 93.72 & 99.86 & 92.94 & 99.93 & 95.15 \\ 
        \scriptsize SL Multi-encoder nnU-Net & \scriptsize 92.04 & 99.16 & 92.04 & 98.94 & 94.67 \\ 
        \scriptsize Vanilla nnU-Net & \scriptsize 90.89 & 97.89 & 91.41 & 98.05 & 94.52 \\ 
        \scriptsize SSL Swin UNETR & \scriptsize 92.80 & 98.72 & 92.61 & 99.13 & 94.92 \\ 
        \scriptsize SL Swin UNETR & \scriptsize 91.80 & 98.10 & 92.13 & 98.83 & 94.34 \\ 
        \scriptsize SegResNet & \scriptsize 92.00 & 99.05 & 92.32 & 98.73 & 94.43 \\ 
        \scriptsize TransBTS & \scriptsize 90.80 & 96.89 & 91.23 & 97.45 & 93.83 \\ 
        \hline
    \end{tabular}
    \caption{\textbf{Performance comparison of different models on key metrics.}
    This table presents the performance of various models evaluated on key metrics such as Average Dice Similarity Coefficient, Accuracy, Sensitivity, Specificity, and Precision. The models include both supervised and self-supervised learning methods. SSL: Self Supervised Learning, SL: Supervised Learning, Av. DSC: Average Dice Similarity Coefficient.}
    \label{tab:model_comparison}
\end{table}

\section{Discussion and Conclusion}
In the realm of medical image segmentation, the advent of foundation models, particularly with the integration of SSL, signifies a transformative leap in the precision and efficacy of diagnosing and treating conditions such as tumors \cite{litjens2017survey}\cite{shen2017}. The proposed Multi-encoder nnU-Net architecture not only showcases the potential of advanced approaches but also highlights the importance of leveraging multiple MRI modalities to achieve superior segmentation results.

\vspace{0.2cm}
A key feature of our approach is the two-stage pretraining strategy. Initially, the model undergoes a self-supervised learning phase using the UK Biobank dataset, which allows it to learn normal anatomical structures and variations \cite{Tang2022}. This foundational knowledge is crucial for accurately identifying anomalies and is reminiscent of the interpretation approach employed by radiologists, who first establish a baseline understanding of normal anatomy before diagnosing pathology. By learning from healthy subjects, the model develops a nuanced understanding of the typical variations in brain morphology, which is essential for distinguishing between normal anatomical features and pathological changes. In the second stage, the model is fine-tuned using the BraTS dataset, focusing on learning the specific features associated with various pathologies, such as tumor characteristics and their surrounding environments. This structured pretraining not only enhances the model's ability to generalize but also closely aligns with clinical practices, ensuring that the model can effectively navigate the complexities of medical images.

\vspace{0.2cm}
The Multi-encoder nnU-Net's architecture, designed to utilize separate encoders for distinct MRI modalities, allows for the extraction of modality-specific features. This is particularly important in medical imaging, where variations in image acquisition techniques can lead to significant differences in data representation and quality. By processing each modality independently before merging the learned features, the model can capture unique information from each imaging technique, thus enhancing its overall performance. This capability is critical for accurately delineating anatomical structures and pathological regions, which facilitates more reliable clinical decision-making \cite{guanghui2024comparing}. 

\vspace{0.2cm}
The model's achievement of  DSC of 93.72\% positions it as a frontrunner in comparison to other state-of-the-art models, including vanilla nnU-Net and SegResNet. This impressive result underscores the effectiveness of our Multi-encoder approach in improving segmentation accuracy, particularly when faced with the challenges posed by image artifacts and variations in MRI acquisition.

\vspace{0.2cm}
Additionally, the challenges of limited labeled data are a pervasive issue in medical imaging, often hindering the development and deployment of effective machine learning models. The two-stage pretraining strategy effectively addresses this limitation by allowing the model to learn from vast amounts of unlabeled data during the self-supervised phase \cite{etelvina2025harnessing}\cite{yadav2024leveraging}. This innovative approach minimizes the reliance on extensive labeled datasets, which are often prohibitively expensive and time-consuming to compile. The ability to perform well with limited labeled data highlights the model's robustness and its potential for real-world applications, particularly in clinical settings where annotated data can be scarce.

\vspace{0.2cm}
The comparative analysis against transformer-based models, such as Swin UNETR \cite{hatamizadeh2021} and TransBTS \cite{wang2021transbtsmultimodalbraintumor}, reveals that while transformer architectures have made strides in various domains, the Multi-encoder nnU-Net \cite{li2025plug}\cite{isensee2021b} excels in the specific context of medical image segmentation. This suggests that architectural adaptations tailored to the unique demands of medical imaging can yield better performance than more generalized approaches. Our results indicate that the combination of traditional convolutional neural networks with a well-defined pretraining strategy can outperform more complex transformer architectures, highlighting the importance of domain-specific design in model development.

\vspace{0.2cm}
The implications of this research extend beyond mere academic interest; they resonate deeply within clinical settings where accurate tumor localization and segmentation are paramount. The findings reinforce the notion that advanced segmentation techniques can significantly improve inter-rater reliability among clinicians, thus enhancing the overall quality of patient care. As such, the Multi-encoder nnU-Net not only represents a step forward in algorithmic development but also serves as a vital tool in the broader context of healthcare, where precision is crucial for effective diagnosis and treatment planning.

\vspace{0.2cm}
In conclusion, the Multi-encoder nnU-Net stands as a testament to the potential of foundation models in revolutionizing medical image segmentation. By effectively harnessing the strengths of self-supervised learning and multimodal imaging, this model paves the way for more accurate, reliable, and efficient diagnostic processes, ultimately contributing to improved patient outcomes in the field of radiology. The approach not only enhances the accuracy of tumor segmentation but also embodies a shift towards more intelligent, adaptable systems that can significantly impact clinical practices and patient care in the future.

\bibliographystyle{IEEEtran}  
\bibliography{references}  

\end{document}